\documentclass{article}

\usepackage{graphicx}
\usepackage{caption}
\usepackage[T1]{fontenc} %computer modern font
\usepackage{authblk}
\usepackage{xcolor}
\usepackage{ulem}

\begin{document}

\title{Real time unsupervised learning of visual stimuli in neuromorphic VLSI systems}

\author[1]{Massimiliano Giulioni\footnote{correspondence to massimiliano.giulioni@iss.infn.it}}
\author[1,2]{Federico Corradi}
\author[1]{Vittorio Dante}
\author[1,3]{Paolo del Giudice}

\affil[1]{Department of Technologies and Health, Istituto Superiore di Sanit\'a, Roma, Italy}
\affil[2]{Institute of Neuroinformatics, University of Z\"urich and ETH Z\"urich, Switzerland}
\affil[3]{National Institute for Nuclear Physics, Rome, Italy}

%correspondence to massimiliano.giulioni@iss.infn.it

\maketitle

\begin{abstract}
Neuromorphic chips embody computational principles operating in the nervous system, into microelectronic devices. In this domain it is important to identify computational primitives that theory and experiments suggest as generic and reusable cognitive elements. One such element is provided by attractor dynamics in recurrent networks. Point attractors are equilibrium states of the dynamics (up to fluctuations), determined by the synaptic structure of the network; a `basin' of attraction comprises all initial states leading to a given attractor upon relaxation, hence making attractor dynamics suitable to implement robust associative memory. The initial network state is dictated by the stimulus, and relaxation to the attractor state implements the retrieval of the corresponding memorized prototypical pattern. In a previous work we demonstrated that a neuromorphic recurrent network of spiking neurons and suitably chosen, fixed synapses supports attractor dynamics. Here we focus on learning: activating on-chip synaptic plasticity and using a theory-driven strategy for choosing network parameters, we show that autonomous learning, following repeated presentation of simple visual stimuli, shapes a synaptic connectivity supporting stimulus-selective attractors. Associative memory develops on chip as the result of the coupled stimulus-driven neural activity and ensuing synaptic dynamics, with no artificial separation between learning and retrieval phases.
\end{abstract}
 
\section*{Introduction}

Since its birthdate in 1989, with the publication of Carver Mead's book \cite{Mead89}, the field of {\em neuromorphic engineering} aims at embodying computational principles operating in the nervous system into analog VLSI electronic devices. In a way, this endeavour may be seen as one modern instance of an over three centuries-long attempt to map forms of intelligent behavior onto a physical substrate reflecting the best technology of the day \cite{Cordeschi}. The additional twist of neuromorphic engineering is a case for a direct mapping of the dynamics of neurons and synapses onto the physics of corresponding analog circuits. 
Initial success was mostly in emulating sensory functions (e.g. visual or auditory perception), and important developments are still ongoing in this area \cite{Liu2013, Liu2010, Liu2015}. However, it soon became clear that the agenda should include serious efforts to emulate, along with such implementations, elements of information processing downstream sensory stages, with the ultimate goal of approaching cognitive functions. 

To make progress in this direction, beyond special-purpose solutions for specific functions, it seems important to identify neural circuitry implementing basic, and hopefully generic, dynamic building blocks, to provide reusable computational primitives, possibly subserving many types of information processing; in fact, this is both a theoretical quest and an item in the agenda of neuromorphic science. Steps in this direction have been taken recently in \cite{Neftci2013}, where `soft winner-take-all' subnetworks provide reliable generic elements to compose finite-states machine capable of context-dependent computation. A review of the electronic circuits involved in such implementations is given in \cite{Chicca2014}.

In recurrent neural populations, synaptic self-excitation can support attractor dynamics, point attractors in the simplest instance on which our approach is based. Point attractors are stable configurations of the network dynamics; from any configuration inside the `basin' of one attractor state, the dynamics brings the network towards that attractor, where it remains (possibly up to fluctuations if noise is present). In a system possessing several point attractor states, the dynamic correspondence between each attractor and its basin implements naturally an associative memory, the initial state within the basin being a metaphor of an initial stimulus, eliciting (even if removed afterwards) the retrieval of an associated prototypical information (memory). For a given network size and connectivity graph, the set of available attractor states is determined by the matrix of synaptic efficacies weighing the links of the graph; {\em Learning} memories is implemented through stimulus-specific changes in the synaptic matrix \cite{Amit89,Wang2008}. 

The attractor-basin correspondence implements dimensionality reduction. Besides, the stimulus-selective self-sustained neural activity following the removal of stimulus that elicited it, can act as a carrier of selective information across time intervals of unconstrained duration, only limited - in the absence of other  intervening stimuli, by the stability of the attractor state against fluctuations \cite{Amit89, Wang2008}.
%First, attractor dynamics implements dimensionality reduction, by mapping sets of stimuli (initial conditions of the network dynamics) onto a
%reduced set of dynamic states to which the dynamical evolution of the network converges (stable patterns of firing rates in the simplest, point attractor instance that will concern us in the present work); this has been suggested to provide a natural substrate for categorizing information. Second, while the attractor to which the network dynamics relaxes is selected by the stimulus, it is self-sustained after the stimulus is removed, and provides a generic mechanism to keep and propagate memory in time.
In a previous paper \cite{Giulioni2012} we demonstrated attractor dynamics in a neuromorphic chip, where synaptic efficacies were chosen and fixed so as to support the desired attractor states.

If attractor dynamics is to be considered as an interesting generic  element of computation and representation for neuromorphic systems, we must address the question of how it can autonomously emerge from the ongoing stimulus-driven neural dynamics and the ensuing synaptic plasticity; this we do in the present work.

To date,  sparse theoretical efforts have been devoted in this direction (see \cite{mongillo2005,Amit95learninginternal,Giudice03}), and to our knowledge this has been never undertaken in a neuromorphic chip.
Here,  in line with our previous papers \cite{Giudice03,GiudiceM01}, and consistently with the above principles, we focus on the autonomous formation of attractor states as associative memories of simple visual objects. 

Our setting is simple, in that our VLSI network learns two relatively simple, and non-overlapped, visual objects. Still, it is complex, in that learning is effected autonomously (that is, without a supervised mechanism to monitor errors and instruct synaptic changes); synapses change under the local (in space and time) guidance of the spiking activities of the neurons they connect, which in turn change their response to stimuli and their average activity because of synaptic modifications. Such a dynamic loop makes the combined dynamics of neurons and synapses during learning quite complex, and controlling it a tricky business; even more so in a neuromorphic analog chip, with the implied heterogeneities, mismatches and the like.

To gain predictive control on the chips' learning dynamics, we first characterize the single-neuron input-output gain function. Then, we use the mean-field theory of recurrent neural networks as a compass to navigate the parameters space of a population of neurons endowed with massive positive feedback and predict attractor states. Finally we measure the rates of change (potentiation or depression) of the Hebbian, stochastic synapses as a function of the pre- and post-synaptic neural activities. These three characterization measures let us choose the correct settings for a successful learning trajectory.  
We then proceed with experiments on the autonomous learning capabilities of the system and
finally, we test the attractor property of the developed internal representations of the learnt stimuli, by checking that when presented with a degraded version of such stimuli the network dynamically reconstructs the complete representation. 

To our knowledge this is the first demonstration of a VLSI neuromorphic system implementing online, autonomous learning.

\section{Results}

\subsection{Neuromorphic Multi-Chip System}
\label{system}
The neuromorphic system (Fig. 1) is composed of a silicon retina \cite{Lichtsteiner_a128} and of two identical reconfigurable neural chips. 
Visual stimuli are displayed on a screen; the silicon retina captures the dynamic contrast of the stimuli and outputs spike sequences. Those spikes are fed into the recurrent learning network implemented on the two neural chips. The network spiking activity is streamed to a PC for analysis. %Arrows, in the schematic view, illustrate the flow of spikes. 

As shown in Fig 1 the recurrent network consists of 196 \emph{Excitatory} and 43 \emph{Inhibitory} neurons physically distributed over the two identical chips. Retina pixels have been grouped into a grid of 14x14 macro-pixels, each one generating convergent output to a single excitatory neuron (see also S.I. Fig. 1).
The recurrent synapses between excitatory neurons are plastic: their efficacy can change depending on the ongoing spiking activity (see below). `Learning' is the stimulus-specific synaptic change induced by the repeated presentation of stimuli and the ensuing neural activity. Excitatory synapses are binary, in the sense that only two state of efficacy are allowed, a \emph{Potentiated} and a \emph{Depressed} one. Learning manifests itself as a sequence of transitions between these two states: as usual we name `LTD', long-term depression the transition from a \emph{Potentiated} to a \emph{Depressed} state, and `LTP', long-term potentiation the \emph{Depressed} $\to$ \emph{Potentiated} transition.

Recurrent synaptic connections are random and sparse (see Fig. 1).
Randomness and sparseness, together with average low values of synaptic efficacy, guarantee low correlations among neuronal activity and, simultaneously, ensure a mean homogeneous input to all neurons. Under these conditions we can approximate the on-chip network behavior with mean-field theory equations (see S.I.) and use predictive theory-inspired tools (i.e. the Effective Transfer Function as explained below) to tune the system parameters. Moreover, the homogeneous connectivity is an ideally unbiased initial condition to test the effect of learning: we expect that synaptic plasticity will cluster the connectivity structure in a stimulus dependent manner.

We provide here a brief descritpion of the neural chips already described in \cite{Giulioni_etal08, Giulioni2009}: the chips are composed each of $128$ Integrate-and-fire (IF) neurons and $16384$ Hebbian plastic bistable reconfigurable synapses. 
Neurons and synapses are designed  as mixed signal analog/digital circuits; the internal dynamics of every element is implemented in continuous-time analog circuitry while communication among neurons relies on digital pulses representing the spikes. The entire chip works asynchronously and in real-time and does not necessitate any clocks. Every neuron and synapse is implemented in silicon with a dedicated circuit, in this way we are able to exploit in parallel all the resources without relying on complex multiplexing schemes. Hence the top-level view of the chip is an ordered matrix of 128x128 synaptic circuits connected to an array of 128 neurons. The drawback of the simplicity of the analog implementation is the unavoidable presence of mismatch deriving from the fabrication process. It causes distributions of the parameters among nominal identical circuits: one of the challenges faced in this work is to gain control over a network of mismatched elements.The analog synaptic circuit implements a plasticity model proposed in  \cite{Fusi00spike-drivensynaptic}, to which we refer the reader for details (see also S.I., section 3); a slightly different version was previously implemented in \cite{Giulioni2009}. 
As already mentioned the synapse is binary, i.e. it has two levels of efficacy  $J_{pot}$ and $J_{dep}$ -- \emph{Potentiated} and \emph{Depressed} which are stable in the absence of pre- and post-synaptic neural activity; `efficacy' is the amount of change in the membrane potential of the post-synaptic neuron per pre-synaptic spike. Plasticity is driven by neural spikes consistently with a rate-based Hebbian paradigm, and the changes of efficacy in each synapse are stochastic because of the irregularity of neuronal spikes in time.
The synaptic connectivity is fully configurable, up to all-to-all connectivity. Each synapse can be set to be excitatory/inhibitory, and can be configured to connect two neurons on the same chip, or on different chips (including the retina), or to accept synthetic spikes from a PC. The communication of spikes from the retina to the neural chips, between the neural chips, and to the PC is based on the parallel asynchronous Address-Event-Representation (AER) \cite{Liu2015, Mahowald:1994:AVS:528544} and it is managed by a custom PCI-AER board \cite{Dante_etal05}, which also implements inter-chips communication. %synaptic connectivity. 

%\sout{We configured the synaptic connectivity on chip to implement a recurrent network of \emph{Excitatory} and \emph{Inhibitory} neurons, which are physically distributed over the two identical neuromorphic chips (see Fig.~\ref{fig:system}1)}.

%\sout{The recurrent synapses between excitatory neurons are plastic: their efficacy (amount of change in the membrane potential of the post-synaptic neuron per pre-synaptic spike) can change depending on the spiking activity of the pre- and post-synaptic neurons. The analog synaptic circuit implements a plasticity model proposed in  \cite{Fusi00spike-drivensynaptic}, to which we refer the reader for details (see also S.I., section 3); a slightly different version was previously implemented in \cite{Giulioni2009}. The synapse is binary (i.e. it has two levels of efficacy  $J_{pot}$ and $J_{dep}$ -- \emph{Potentiated} and \emph{Depressed}); plasticity is driven by neural spikes consistently with a rate-based Hebbian paradigm, and the changes of efficacy in each synapse are stochastic because of the irregularity of neuronal spikes in time. `Learning' is the stimulus-specific synaptic change induced by the repeated presentation of stimuli and the ensuing neural activity. It manifests itself as a sequence of transitions between the two allowed synaptic efficacy states (\emph{Potentiated} $\to$ \emph{Depressed}: `LTD', long-term depression; \emph{Depressed} $\to$ \emph{Potentiated}: `LTP', long-term potentiation). }
%\sout{For the original motivation and theoretical setting see \cite{Amit89}}.
%%%%%%%%% FIGURE SYSTEM HERE
\begin{figure}
\begin{center}
\centerline{\includegraphics[width=12cm]{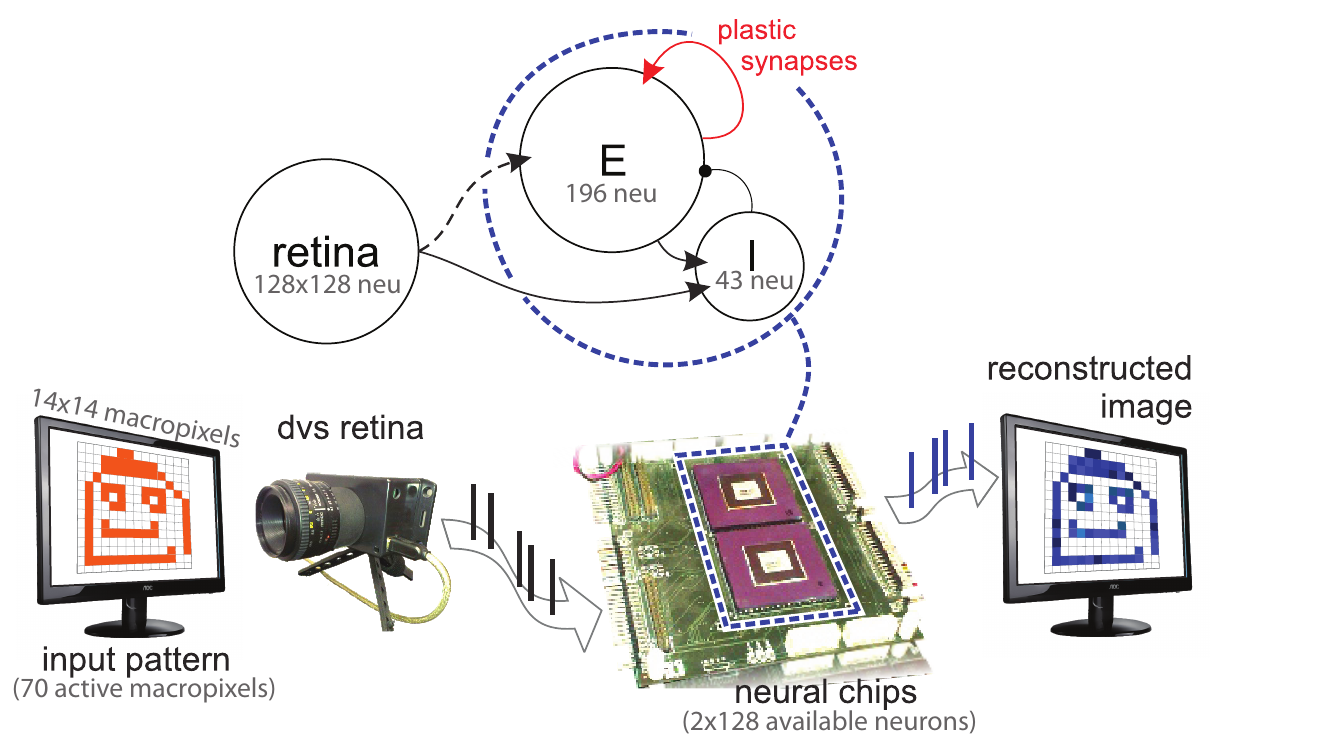}}
\caption{\textbf{The neuromorphic system}. A visual stimulus is shown on a screen in front of the retina chip. The retina, a matrix of $128x128$ pixels, outputs spikes to two neural chips configured to host a recurrent network with a population $E$ of $196$ excitatory neurons and an inhibitory population $I$ of $43$ neurons (network's architecture sketched on top). Sparse connections, shown with a solid line, are generated randomly: connectivity levels, i.e. the probability of synaptic contact, are $0.25$ for $E$ recurrent connection, $0.5$ for $I$ to $E$ connection, $0.3$ for $E$ to $I$ and 0.02 from retina to $I$ connection. The retina field of view is divided into $14x14$ macropixels, each projecting onto a single neuron of $E$. The network's spiking activity is monitored by a PC.}
\end{center}
\end{figure}\label{fig:system}

%\sout{In order to map the retina output onto the input of the neural chips, retina pixels have been grouped into macro-pixels, each one generating convergent input to a single excitatory neuron (see S.I. Fig. 1).}
%Moreover, the homogeneous connectivity is an ideally unbiased initial condition to test the effects of learning, which we expect to shape the connectivity building a structure dependent on the input stimuli.}

% Retina input is noisy, due to the intrinsic noisy activity of the low-power analogue pixels.  
\subsection{Theory-inspired tools to control the neuromorphic system}
As remarked in the Introduction, the successful autonomous, unsupervised development of associative representations of stimuli in the form of attractors of the network's dynamics can be challenged in many ways by the interplay between neural and synaptic dynamics. Because of this, and given the large network parameter space, setting neural and synaptic parameters by trial and error is not a viable route.
To gain control over the system we proceed in three steps: 1) we tune neuronal parameters by measuring the input-output response function of a single-neuron 2) then, to set the synaptic efficacy values,  we characterize the response of a recurrent sub-population of neurons connected by non-plastic synapses, 3) finally, to choose the settings for the plasticity circuitry, we measure the LTD and LTP probability of the on-chip synapses when subjected to controlled neuronal activity. For the first and third measures we developed ad-hoc experimental procedures (described respectively in the second section of the S.I. and in the next section). To study the response of the recurrent neuronal sub-populations we faced challenges related to the characterization of a system endowed with a massive positive feedback and embedded in a larger network. Relying on predictions based on theoretical models and mapping the derived parameters onto the hardware is not viable: on one side it requires lengthy calibration procedures and on the other side, anyhow, it is prone to fail due to unavoidable differences between the models' assumptions and analog circuital behavior, which easily became critical in presence of a massive positive feedback. 
%\sout{On the other hand, relying on theoretical predictions based on models, and simply mapping the derived parameter values onto the chip not only requires lengthy calibration procedures, but is prone to fail due to several factors that violate in the chip hypotheses needed for the validity of the theoretical predictions.}
%\sout{As an alternative, after characterizing the input-output transfer function of the silicon neuron as described in the SI, section 2,} 
As an alternative, we used the method introduced and validated in \cite{Giulioni2012} to identify regions in the parameter space compatible with the desired network behavior: the coexistence of (two, in our case) stable, stimulus-selective collective attractor states; in the SI, section 2, we briefly summarize the procedure, and describe the results for the present system. The method instantiates the dimensional reduction of the mean-field theory for a multi-population network (proposed in \cite{MascaroAmit1999}) in a self-consistent procedure on chip, allowing us to estimate the `Effective Transfer Function' (ETF) of a population of neurons of interest (one selective excitatory population in our case). From an engineering point of view the ETF can be seen as the open-loop transfer function of the sub-population of interest, taking into account the feedback provided by the rest of the network. As detailed in the S.I. the ETF allows us to predict, within certain approximations, the mean firing rate of attractor states of a single sub-population for different levels of its recurrent synaptic potentiation. Since learning implies selective changes in the ratio of the potentiated/depressed synapses in the network, the study of ETF for the sub-populations which will be affected by the stimuli offers a predictive tool for expected learning histories.

\subsection{Synaptic plasticity}
\label{sec:synapticdyn}
The analysis based on the ETF gives an estimate of the expected changes in the average firing rates of the network's attractor states, as the ratio of potentiated/depressed synapses changes as a result of learning. In turn, expected rates of synaptic changes evolve depending on changes in the network's populations average activities. It is therefore relevant to derive an estimate of the LTP and LTD transition probabilities as a function of the pre- and post-synaptic firing rates ($\nu_{\mathrm{pre}}, \nu_{\mathrm{post}}$); the procedure is described in Materials and Methods section.
%The range of firing rates induced by the retina stimuli in the excitatory neurons of the recurrent network (predicted by the ETF, see S.I.) should match the sensitivity of the synapse in terms of LTP and LTD probabilities.
%We therefore performed a `high-level' characterization of the synaptic circuit, by systematically estimating the LTP and LTD probabilities as functions of the pre- and post-synaptic firing rates ($\nu_{\mathrm{pre}}, \nu_{\mathrm{post}}$, respectively), shown in Fig.~{\ref{fig:LTP_LTD}}. 

%%%%%%%%%%%%%% FIGURE 2 LTP/LTD HERE
\begin{figure}
\begin{center}
\centerline{\includegraphics[width=10cm]{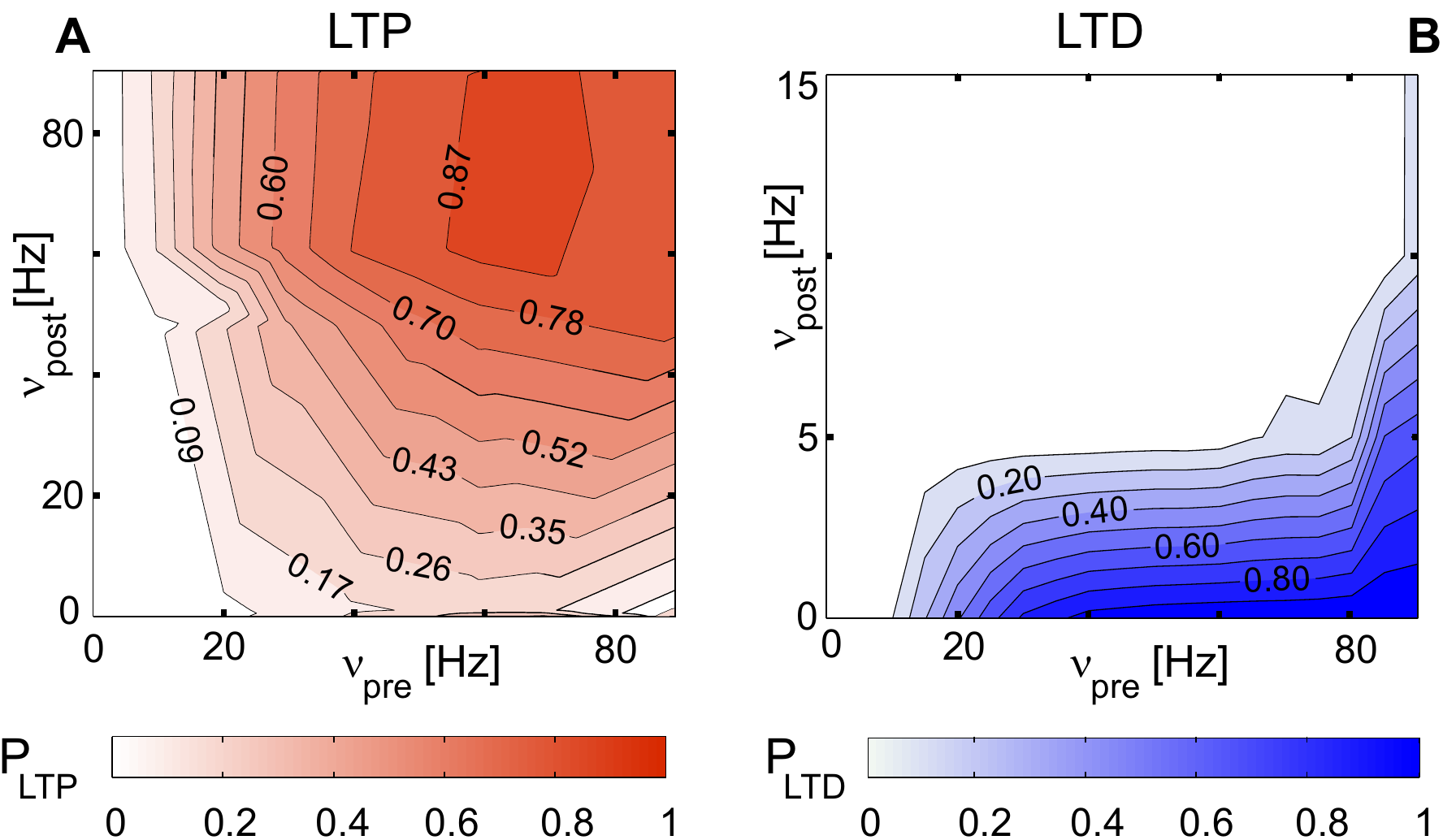}}
\caption{~\textbf{Long Term Potentiation (LTP) and Depression (LTD) probability}. On the x axis we report the pre-synaptic neuronal mean firing rate $\nu_{\mathrm{pre}}$, on the y axis the post-synaptic neuronal firing rate $\nu_{\mathrm{post}}$. Details on the experimental procedure are reported in methods}
\end{center}
\end{figure}\label{fig:LTP_LTD}

The contour plots in Fig.2, panel B confirm that the silicon synapse implements, as desired, Hebbian LTP, (its probability increases as $\nu_{\mathrm{pre}}$ and $\nu_{\mathrm{post}}$ both increase) and heterosynaptic LTD (i.e., synaptic depression probability increases with increasing $\nu_{\mathrm{pre}}$, and only occurs for low $\nu_{\mathrm{post}}$).

The joint information from the analysis of the ETF and the LTP/LTD transition probabilities allowed us to approximately predict the `working point' of the plastic synapses at successive stages of learning (the prediction is expected to be more reliable for slow learning, such that the network evolves through quasi-equilibrium states, which are difficult to obtain for a small and heterogeneous network like the one on chip). Such knowledge allows to promote `learning trajectories' with balanced LTP/LTD changes, which is another important stability factor \cite{Giudice03}.
%
%
%we could reliably predict the stationary firing rates, also in response to stimulation, of the neural population of interest for different fractions of potentiated synapses; in turn, this allowed us to approximately predict the `working point' of the plastic synapses during learning, assuming that learning itself, i.e. the process of synaptic changes induced by stimuli, was slow enough to allow considering the network evolving through `quasi-equilibrium' states (actually this was checked to be approximately true, up to the very earliest stages of learning).

\subsection{Autonomous learning: forming stimulus-selective attractor states}

Two visual stimuli were repeatedly presented on a LCD screen, acquired in real time by the silicon retina, and mapped onto the recurrent network distributed on two neural chips; our goal was to achieve autonomous associative learning leading to the formation of stimulus-selective attractor states as internal representations (`memories') of the stimuli. Hence, after learning, for every stimulus we expect a specific network response that should persist even after the removal of the stimulus.

Visual stimuli, (a `happy' and a `sad' face shown in panel A of Fig.3), are orthogonal (zero overlap) and their coding level (fraction of activated macro-pixels, see S.I.) is fixed at about $1/3$; Hence each stimulus activates about 5460 retina neurons and, correspondingly, about 65 excitatory neurons. The implications of this for more realistic situations are discussed in the Discussion section.

Learning proceeds as a sequence of transitions between the $J_{pot}$ and $J_{dep}$ synaptic states, depending on the activity of pre- and post-synaptic neurons induced by stimuli, as explained below. Notice that no explicit control is imposed on synaptic dynamics depending on the network being stimulated or not, and no distinction is made between `learning' and `retrieval' phases; the network just evolves based on the incoming flow of stimuli, and its own feedback: our system's learning is completely autonomous and does not require any supervision. 

According to the Hebbian learning that synapses implement (as from Fig.2), each stimulus presentation is expected to provoke changes in a fraction of synapses as follows (we remind that learning is stochastic, to the extent that neural activities are): LTP in synapses connecting neurons activated at high rates by the same stimulus; LTD in synapses connecting neurons activated by different stimuli, or connecting neurons activated by stimuli to neurons never activated by any stimuli (we named them the `background' neurons); statistically little or no changes in  synapses connecting pairs of `background' neurons. 

In the above scenario, because of autonomous learning, excitatory neurons get partitioned in three populations (two selective to stimuli, and the background), which both under stimulation and in the absence of it show an evolving pattern of relatively stable firing rates; as remarked in the Introduction, during this unsupervised evolution the network could well drift to undifferentiated high activity or quiescence states. Balance between LTP and LTD is essential to guarantee a successful learning trajectory \cite{GiudiceM01}. LTP should eventually grow high enough to support stable selective attractor states; however, if this is not counterbalanced by LTD along the way, and especially in the early stages of learning, the learning trajectory can easily lead the network to globally unstable states; knowledge of the LTP/LTD probabilities of Fig.2 is important to obtain robust learning trajectories.
This is where the predictive power of the ETF, and knowledge LTP/LTD curves, come into play, hinting at safe paths in the large parameter space. 

Our main results are described in Fig.3, which illustrates a typical successful `learning history', and in Fig.4, which describes the underlying evolving micro-structure of the synaptic matrix. 
%\st{The top row shows the sequence of stimuli (from now on denoted by `happy face', `sad face') displayed on the screen}; \st{we plot the corresponding single-trial firing rates of the network's sub-populations (the two excitatory ones affected by the two stimuli are platted in red -- happy and green -- sad). The bottom panel shows a two-dimensional representation of the network activity during the inter-stimulus intervals, constructed from the map of the macro-pixels to the neurons.}%\massi{The sequence of visual stimuli shown in Panel A induces the firing rate of the network reported in panel B and re-plotted, for the inter-stimulus intervals, in two-dimensional images in the Panel C}

Panels {\tt A,B,C} of Fig.3 describe respectively the sequence of stimuli, the average firing rates of the two selective populations and a two-dimensional representation of the network's output to match the representation of stimuli.
It is seen that: 1) initially, each stimulus provokes a response in its target neural population (which, through the induced inhibitory activation essentially silence the other populations, up to small noise); this activity is rapidly extinguished when the stimulus is removed, and no noticeable activity is present during the inter-stimulus interval 2) during each stimulation, a fraction of synapses changes, according to the (stochastic) scheme already explained, and this is reflected in the slow increase in the firing rates under stimulation 3) after many repeated stimulations, the build-up of synaptic self-excitation determines the appearance of a high-activity (meta)stable state for each of the two selective sub-populations (the equivalent of the ETF passing from 1 to 2 stable fixed points in Fig 3 in S.I.), and after the stimulus is released, the corresponding selective sub-population stays in a self-sustained state of elevated activity: the attractor state propagating the memory of the last stimulus across the inter-stimulus interval; sometimes this reverberant state is destabilized by the next incoming stimulus, while in other cases it decays spontaneously, due to finite-size fluctuation. Correspondingly, in a phase of mature learning, the output the of the network during the inter-stimulus interval well matches the input stimuli.

%%%%%%%%%%%%% FIGURE 3 HERE
\begin{figure}
\begin{center}
\centerline{\includegraphics[width=17.8cm]{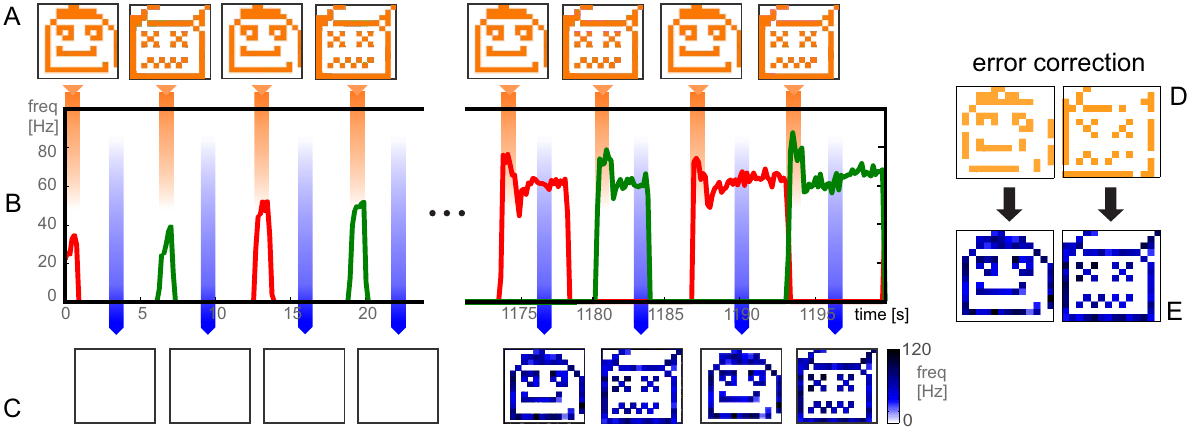}}
\caption{~\textbf{Learning Dynamics.}~\textbf{A}: Sequence of input visual stimuli each active for 1 sec. as highlighted by the orange bands. \textbf{B}: single-trial firing rates of the excitatory sub-populations responding to the \emph{happy} (red) and \emph{sad} (green) faces. \textbf{C}: system output, i.e. network activity during the inter-stimulus intervals shown as a two-dimensional image, constructed from the map of the macro-pixels to the neurons. \textbf{D}: degraded version of input stimuli and \textbf{E} corresponding network output.}
\end{center}
\end{figure}\label{fig:learningpsth}

%In the initial condition $5 \%$  of the recurrent excitatory synapses are set in the potentiated state.

A sample history of the synaptic changes underlying the learning history of Fig.3 is given in Fig.4, Panels A, B. Panel A illustrates the time course of the fraction of synapses in the potentiated state, for the different synaptic groups (identified by colour); synapses are inspected every 2 presentations of the same stimulus (4 stimulations in the alternate sequence). Consistently with expectation:  synapses connecting neurons affected by the same stimulus get potentiated, actually approaching saturation, such that the average potentiation level remain essentially stable for $t>300$s; LTD is visible in the initial stage of learning; synapses connecting neurons in the background stay essentially unaffected.

Panel B shows the Hamming distance between the network synaptic matrices sampled as in panel A. %, for the whole learning history, subdivided in the same synaptic sub-groups as in the A panel. 
We see here that the seemingly steady situation reached in panel A for the selective `sad to sad' and `happy to happy' synapses actually conceals a dynamic balance; for this to happen, again, LTP and LTD probabilities should be properly balanced.
The evolution of the synaptic matrix is further illustrated in panels C--E of Fig.4. %In each panel we represent the matrix of all recurrent synapses between excitatory neurons; rows (columns) in the matrix contain the index of pre-synaptic (post-synaptic) neuron, ordered such that indices of neurons stimulated by each stimulus, as well as those of background neurons, are continuous (identified by colors, consistent with panels A and B). Black (grey) dots indicate synapses in the potentiated (depressed) state.
The pictures are, from top to bottom, snapshots of the synaptic matrix taken at the beginning of learning, after 30 seconds, (2 presentations of each stimulus), and after 300 seconds (40 presentations).

The snapshot in panel C just reflects the random choice of $5 \%$ potentiated synapses as the initial condition of the network; in panel D we see that the synaptic matrix getting the expected structure, for the chosen index labeling: two blocks of mainly potentiated synapses (LTP), corresponding to synapses connecting neurons responsive to the same stimulus (remember that the stimuli are orthogonal, hence the non-overlapping `potentiated squares'); `whitened' blocks of depressed synapses (LTD) connecting neurons responsive to different stimuli, or those connecting neurons responsive to either stimuli to background neurons; one strip on the extreme right, of synapses connecting neurons in the background, which stay essentially unaffected. Such features get further sharpened in the mature learning phase of panel E.

\subsection{Error Correction properties}
\label{error-correction}
In order to check that the developed selective, self-sustaining states of elevated activity are indeed attractors of the network dynamics, 1) we observe their persistence after removal of the stimulus and 2) we check that the network is able to perform {\it error correction} of a degraded stimulus: in a mature stage of learning, for initial conditions (stimuli) close to one of the `memories', the network spontaneously relaxes to the corresponding attractor state. %\sout{in other words, that the network is able to perform {\it error correction} of a degraded stimulus.}

This we show in panel D of Fig 3, in which the stimulus is degraded by removing $20\%$ of its active pixels.  The network quickly reconstructs the complete, learnt memory, thanks to the selective feedback. We remark that systematically exploring the error-correction ability, by varying the amount of degradation of the stimuli, would allow to give an estimate of the {\it basin of attraction} of the attractor states. Indeed, this we showed in \cite{Giulioni2012} (in which the selective synaptic structure was imposed and not self-generated).
%In our case the structure of the basins is trivial because of the chosen orthogonality of the stimuli: indeed, even removing $90 \%$ of the pixels defining the stimulus, the network correctly reconstructs its complete memory. However, we remark here that, besides the imposed degradation of the input stimuli, a form of uncontrollable degradation is anyway present because of the noisy response of the silicon retina across its pixels, which is enough to induce a large distribution of activity at the level of the macro-pixels defined in Section {\tt Neuromorphic Multi-Chip System}.

%In the Discussion we briefly elaborate on the implications of relaxing the orthogonality constrain.

\section{Discussion}

We demonstrated a neuromorphic network of spiking neurons and plastic, Hebbian, spike-driven synapses which autonomously develops attractor representations of real visual stimuli acquired by a silicon retina. Attractor dynamics results in stimulus-selective, elevated activity after removal of the stimulus, and error correction properties. Detailed inspiration from mean-field theory is used to implement on chip a search strategy in the large parameter space of the network, which works at the level of the network's input-output gain function, without having to delve into the single circuits level.

%%%%%%%%%%%%%%%%%%% FIGURE 4 HERE
\begin{figure}
\begin{center}
\centerline{\includegraphics[width=12cm]{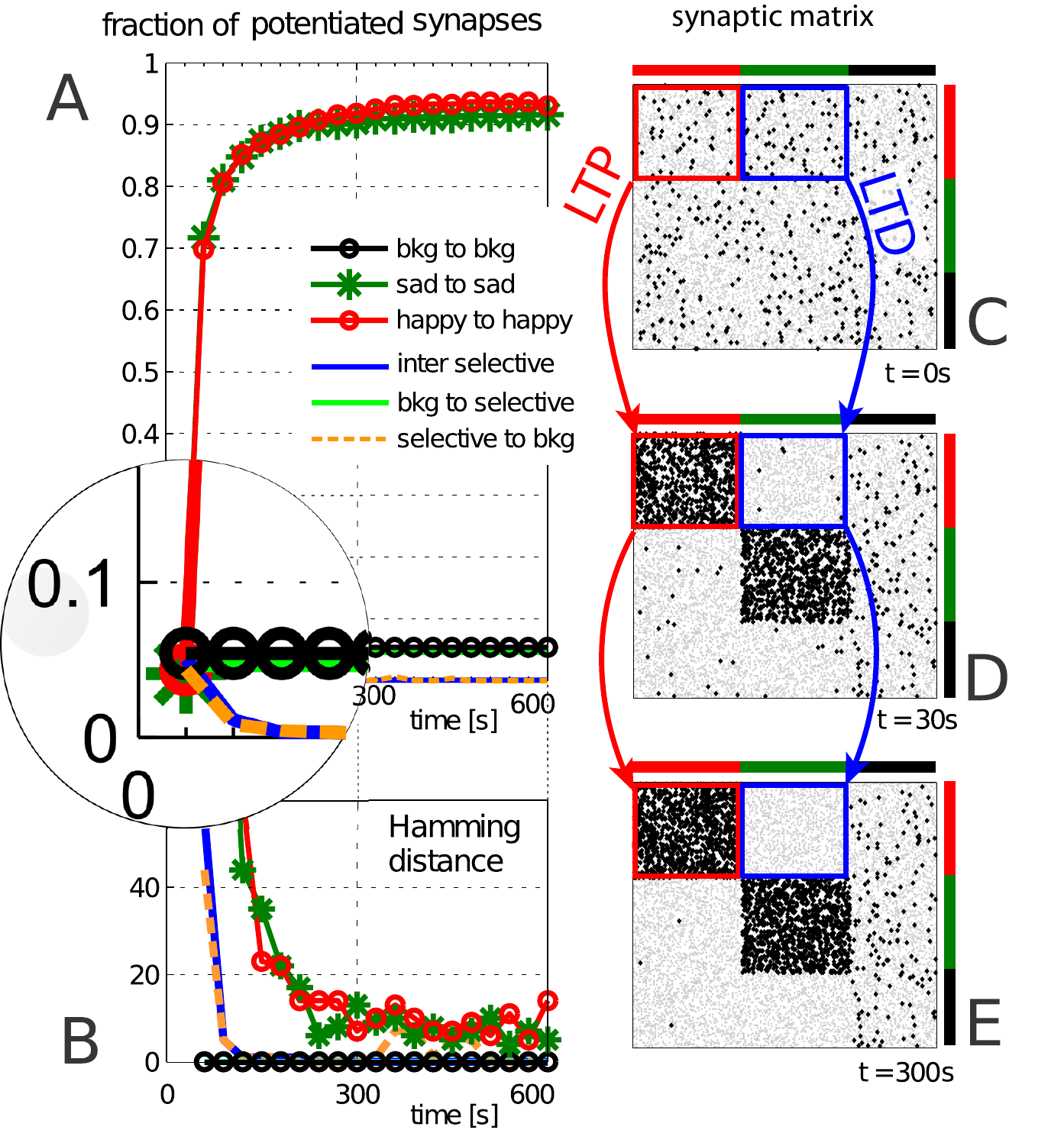}}
\caption{~\textbf{Synaptic evolution.}~\textbf{A}: Fraction of potentiated synapses. Traces named \textit{inter-selective}, \textit{selective to bkg} and \textit{bkg to selective} refer respectively to the average fraction of potentiated synapses in the connections between the two selective sub-populations (happy and sad), from the two selective sub-populations to the background, and from the background to both happy and sad.  \textbf{B}: Hamming distance over time between two successive recordings of the synaptic matrix. \textbf{B C D}:  snapshots of the excitatory recurrent connection: gray (black) dots are depressed (potentiated) synapses between the pre-synaptic neurons (plotted on the x-axis) and post-synaptic ones (on the y-axis). Neurons responding to the same stimulus are grouped together. In red we highlight the group of synapses connecting neurons strongly reacting to the happy face. They get potentiated during learning. The group of synapses highlighted in blue are those connecting sad to happy population: as expected they get depressed during the initial learning stages.}
\end{center}
\end{figure}\label{fig:syn}

Within the limitations of the chosen scenario, which we discuss in the following, what we achieved takes a step along a most needed line of development, if neuromorphic systems are ultimately to act as autonomous and adaptive dynamic systems interacting in real time with their environment (for which, of course, much more than just building associative representations will be needed). 

In view of further progress, some of the simplifying choices we made, and some limitations must be discussed and put in perspective.

First, it would be tempting to compare the performance of our neuromorphic network with theoretical predictions about the limit of capacity, i.e. the maximum number of networks configurations that can be embedded as retrievable memories. Hebbian dynamic learning with bounded (two-states in our case) synapses entail a capacity scaling as the logarithm of the number of synapses \cite{Fusi2002} (as opposed to the linear scaling of the Hopfield network \cite{Hopfield1982,Amit89} or its equivalent realization with spiking neurons \cite{Curti:2004:MFC:1119696.1119702}); this limit can be partially overcome by making learning stochastic (as in our case) or lowering the coding level (see \cite{FusiAbbott2007} for a discussion and further developments). However, a quantitative comparison with the theoretical estimates is affected by several factors (for the chip and largely for simulations as well), e.g. the fact that the probability of synaptic changes is not really constant along the learning history; that mismatch in the synaptic circuits effectively creates - at any stage of learning - a distribution of probabilities of synaptic changes across the network (notice that we do not compensate for mismatch modulating the connectivity among neurons as suggested in \cite{Neftci_Indiveri10}); that finite-size effects influence synaptic dynamics through the distribution of firing rates they induce \cite{Giudice03}.

Empirically, the memory capacity of our small VLSI network is three patterns, beyond which learning becomes unstable; of course the small size of the network prevented us from checking how the capacity scales with the number of synapses. %On the other hand we remark that attractor dynamics is quite robust to both quenched noise (i.e. mismatch) and to dynamic finite-size noise.
From the hardware point of view, achieving higher capacity with larger network hits at a scalability issue. Designing new chips that embed more neurons and synapses would provide a limited option for scaling up (unless radical changes in the implementation technology are considered - e.g. memristors). The other natural approach would be to combine many neural chips; this would face challenges related to the bandwidth for inter-chip communication, that would need to ensure reliable real-time AER-like spike delivery.

Second, we choose orthogonal stimuli (also with the same coding level) to be learnt by the network, which clearly facilitates learning by minimizing the interference induced on synaptic changes by different stimuli. A natural question then arises, as to the implication of including more naturalistic stimuli with arbitrary overlaps (and possibly coding levels). 
As for overlaps, a computationally effective, and biologically motivated, modification of synaptic dynamics was proposed in \cite{Brader2007}; for the same stochastic, bistable Hebbian synaptic model here implemented, a regulatory mechanism was there introduced, preventing further potentiation or depression of a synapse when the post-synaptic neuron was recently highly active or poorly active, respectively. 
%The rationale for the prescription is to de-emphasize the coherent synaptic potentiation that would result for synapses affected by the common part of different patterns and that would ultimately spoil the ability to distinguish the patterns. 
Such modified `stop-learning' synapses were implemented and successfully demonstrated in neuromorphic chips \cite{Giulioni2009, Mitra_etal09}, and would provide a good option to make the development of attractor representations more robust to the spatial structure of the stimuli. 
Allowing for different coding levels is instead essentially a matter of size of the network, allowing, even for low coding level, for a good averaging of synaptic mismatches across the dendritic tree of each neuron. 

%{\em The stability, and reproducibility, of learning histories is affected by finite-size effects (see \cite{GiudiceM01} for an analysis of such effects). Of those, some are due to the finite number of neurons, inducing fluctuation in the network firing rate, and \massi{some other} to the quenched randomness of the specific realizations of the sparse synaptic connectivity, at parity of its average properties; those effects  are inherent in the architectural choice, and would be present even in a numerical simulation; in general, large distribution of firing rates can distort the synaptic dynamics. On top of that, our silicon instantiation of the neural network also suffers from circuital mismatches and temperature dependence; the former is another source of quenched noise, while the second, within certain limits (about $\pm 1$ \textcelsius), just speeds-up or slows-down the pace of learning; if temperature variations are too large the correct working of the chips is lost.  }

Putting our work in the context of previously published work, the attempt to embed learning capabilities in hardware devices is of course not new in general, starting from the supervised `adaptive pattern classification machine' described by Widrow and Hoff \cite{Widrwo1960} in the 60s.

%, of a  `adaptive pattern classification machine' made out of discrete components which learns to classify linearly separable patterns exploiting a perceptron-like architecture and supervised learning. 
Not exhaustively: one of the first steps towards learning microchips was taken in the 80s by Alspector and colleagues \cite{Alspector1988}, who demonstrated unsupervised learning in a feed-forward network (for a broad review of pioneering works see \cite{Cauwenberghs1998}). More recently, \cite{Hafliger2007} demonstrates a feedforward architecture to classify patterns of mean firing rates imposed by synthetic stimuli. A complex neuromorphic chain including visual sensing and processing (convolution filters) was described described in \cite{caviar2009}. 

Massive efforts have been devoted to develop various kinds of plastic synaptic circuits (using standard CMOS 
\cite{Hafliger2003,Gordon2002,Liu2008,Ramakrishnan2011,Petit2004,
Schemmel2006,Bamford2012,Rachmuth2011,Indiveri2006,Indiveri2010,Brink2012}
or with new memristor devices 
\cite{suri2012,mayr2012,serrano2013,gelencser2012}; for a recent review of technologies employed for the purpose see \cite{Kuzum2013}).

Still, very few works demonstrated learning in neuromorphic hardware at the network level. As mentioned above, Hebbian stop-learning synapses \cite{Brader2007} have been used in pure feed-forward networks \cite{Giulioni2009, Mitra_etal09} trained in a semi-supervised way to discriminate among syntethic patterns of mean firing rates, while \cite{Arthur2006} demonstrates how Spike-Timing-Dependent-Plasticity increases synchronicity in a network with local connectivity among neighboring neurons. 

To our knowledge, learning synthetic stimuli in a spiking recurrent network with massive feedback has been dealt with only off-chip in \cite{SeoBLPEMRTCMF11}, reporting a digital implementation. 

In the domain of recurrent spiking networks implementing associative memories this work takes a significant step towards autonomous operation in naturalistic conditions by removing - as we did - the artificial separation between a `learning phase', in which pre-computed synapses are `downloaded' to the chip, and a `retrieval phase', in which the associative memory is tested with frozen synapses. %\sout{takes a significant step towards the conditions of autonomous operation in naturalistic conditions.}

Finally, concerning the prospective interest of point attractor networks as `reuseble building blocks' of neuromorphic systems (see Introduction) we would like to remark that, in the face of the rich repertoire of dynamic states exhibited by the brain (and neural networks), attempts are under way to bridge such elementary point attractor dynamics and the complexity of neural dynamics over multiple time scales (see \cite{BraunMattia2010} and references therein), and the domain of application of the attractor concept has gradually evolved to include models of working memory, information integration, decision making, multi-stable perception. This provides, we believe, a prospective rich context for the implementation reported here.

\section{Materials and Methods}
\subsection{Procedure for estimating LTP and LTD probabilities}
We choose 64 excitatory neurons, and for each one we set up 128 input synapses, all configured as `AER synapses', of  which 64 are configured to be non-plastic (efficacy set to 0.05), the remaining 64 were plastic, with efficacy set to zero (in this way they do not affect the neuron's firing, but the synapse can still perform transitions between its two binary states). In this way, firing $\nu_{post}$ of each neuron is driven by input from the 64 non-plastic AER synapses, receiving on their pre-synaptc terminal synthetic spike trains through the PCI-AER board; the plastic AER synapses are pre-synaptically driven by synthetic spike trains with average rate  $\nu_{pre}$. To explore the ($\nu_{pre}, \nu_{post}$) plane, for each $\nu_{post}$ (i.e. for each average rate of external spikes to the non-plastic synapses), $\nu_{pre}$ (the average rate of external spikes to the plastic synapses ) is varied. For each ($\nu_{pre}, \nu_{post}$) pair, the initial state of the synapse (i.e. the value of the internal variable $X(0)$) is set (high, to measure LTD, low, to measure LTP), and its binary state is checked after 1 sec.

\bibliographystyle{naturemag}

\section*{Acknowledgement} This work was partially supported by the EU FET project Coronet. We are indebted to Tobi Delbruck for having made available to us early prototypes of the silicon retina, and for his assistance. We gratefully thank Maurizio Mattia for his precious contribution of ideas and suggestions along the way, and Stefano Fusi for a critical reading of the manuscript.

\section*{Author contributions} M.G. and F.C. wrote the software, performed the experiments, analysed the results and wrote the paper. V.D. designed the test system and provided the firmware. P.d.G. supervised the entire work and contributed to data analysis and paper writing. All authors reviewed the manuscript.

\section*{Correspondence} Correspondence and requests for materials
should be addressed to Massimiliano Giulioni~(email: massimiliano.giulioni@iss.infn.it).

\end{document}